\documentclass[conference]{IEEEtran}
\usepackage[T1]{fontenc}
\usepackage[utf8]{inputenc}
\usepackage[normalem]{ulem}
\usepackage{cite}
\usepackage{amsmath,amssymb,amsfonts}
\usepackage{algorithm}
\usepackage{algorithmic}
\usepackage{graphicx}
\usepackage{textcomp}
\usepackage{xcolor}
\usepackage{url}
\usepackage{bbm}
\usepackage{hyperref}
\usepackage{multirow}

\newcommand{\probP}{\text{I\kern-0.15em P}}
\def\BibTeX{{\rm B\kern-.05em{\sc i\kern-.025em b}\kern-.08em
    T\kern-.1667em\lower.7ex\hbox{E}\kern-.125emX}}

\makeatletter 
\newcommand{\linebreakand}{%
  \end{@IEEEauthorhalign}
  \hfill\mbox{}\par
  \mbox{}\hfill\begin{@IEEEauthorhalign}
}
\makeatother 

\begin{document}

\title{Unsupervised Feature Construction for Anomaly Detection in Time Series - An Evaluation}


 \author{
 \IEEEauthorblockN{Marine Hamon}
 \IEEEauthorblockA{\textit{Orange Innovation}\\
 Lannion, France \\
 marinehamon30@gmail.com}
 \and
 \IEEEauthorblockN{Vincent Lemaire}
 \IEEEauthorblockA{\textit{Orange Innovation}\\
 Lannion, France \\
 vincent.lemaire@orange.com}
 \and
 \IEEEauthorblockN{Nour Eddine Yassine Nair-Benrekia}
 \IEEEauthorblockA{\textit{Orange Innovation}\\
 Paris, France \\
 noureddineyassine.nairbenrekia@orange.com}
 \and
 \linebreakand
 \IEEEauthorblockN{Samuel Berlemont}
 \IEEEauthorblockA{\textit{Orange Innovation} \\
 Meylan, France \\
 samuel.berlemont@orange.com}
 \and
 \IEEEauthorblockN{Julien Cumin}
 \IEEEauthorblockA{\textit{Orange Innovation} \\
 Meylan, France \\
 julien1.cumin@orange.com}
 }

\maketitle
\begin{abstract}
To detect anomalies with precision and without prior knowledge in time series, is it better to build a detector from the initial temporal representation, or to compute a new, tabular representation using an existing automatic variable construction library? In this article, we address this question by conducting an in-depth experimental study for two popular detectors : Isolation Forest and Local Outlier Factor. The results, obtained from experiments on five different datasets, show that the new representation, computed using the \textit{tsfresh} library, allows Isolation Forest to significantly improve its performance.
\end{abstract}



\section{Introduction}
\label{sec-introduction}

The literature of time series deals with various learning tasks such as forecasting, clustering and classification. In this article, we address the problem of time series anomaly detection (TSAD) \cite{boniol-slides,outlier-anomaly_detection,adbench,comprehensive_survey_def_anomaly,Boniol-decade-tsad}, specifically for univariate time series. We denote $\tau_i = \left\langle (t_1, x_1),\ldots, (t_{m},x_{m})\right\rangle$ a univariate time series, where $x_k$ is the value of the series at time $t_k$. In this article, the objective is to predict whether the pair $(t_i, x_i)$ corresponds to an anomaly (point-wise detection).

Over the past few years, a consensus has emerged within the community that transforming time series from the temporal domain to an alternative representation space is one of the most effective ways of improving model accuracy. This has been observed in classification \cite{classif,cocalite,FreshPRINCE}, early classification \cite{EDM} and, to a lesser extent, in anomaly detection in a few very specific publications focused on particular use cases\cite{features_engineering_PCA,features_engineering_DBSCAN}.

This recent research work has motivated the study we present in this paper. We build on the idea of changing the representation space and then applying ``usual'' anomaly detectors designed for tabular data \cite{classif}. The question we address here is: can we achieve better anomaly detection performance in the computed feature space, or is it better to stay in the original temporal representation?

The rest of this article is organized as follows: section \ref{sec-concepts} presents the background and key concepts used in this article, so that it can be properly positioned in the very large literature of anomaly detection for time series. Section \ref{sec-pipeline} presents the proposed processing pipeline. Section \ref{sec-protocole} presents the experimental protocol. Detailed results are then presented in section \ref{sec-resultats} before concluding in the final section.


\section{Context and Concepts Used}
\label{sec-concepts}
To allow the reader to correctly position the work carried out in this article, we outline below the main themes present in the literature of anomaly detection for time series \cite{boniol-slides,Boniol-decade-tsad} and explain our positioning. As the topic of anomaly detection in time series is very widely covered in the literature, we do not claim to be exhaustive but rather aim to be factual for the purposes of this study.

\subsection{Types of Anomalies}

The literature distinguishes at least three main types of anomalies\cite{Def_anomaly_Chandola}: 
\begin{itemize}
    \item (i) \textbf{point anomalies} : for example, an unusually high financial transaction in relation to a customer's transaction history.
    \item (ii) \textbf{collective anomalies} (in the sense of a succession of correlated punctual anomalies): for example, a sudden drop in traffic on a website, due for example to a server failure or a denial-of-service attack.
    \item (iii) \textbf{contextual anomalies}: for example, an abnormal increase in electricity consumption in a given region, due for example to a snowstorm or an exceptional heat wave.
\end{itemize}

Given the objective of our study, we are placing ourselves in the most general case possible. We make the effort of being as “agnostic” as possible through the following assumptions. (i) We have no feedback loop from an expert user to contextualize the observed data (ii), to adjust the parameters of the methods (iii), or to adjust the sliding window size (see below). (iv) We also do not try to distinguish punctual anomalies from collective ones.

\subsection{Online or Offline Analysis}

For anomaly detection in time series, it is important to distinguish between online detection and offline analysis. Online detection refers to the real-time detection of anomalies as they occur. It is often used in real-time monitoring systems where quick response is essential to prevent undesirable events. Offline analysis, on the other hand, focuses on the retrospective examination of data to identify anomalies after the fact. It is generally used for post-hoc analysis, understanding the root causes of anomalies and improving online detection systems. In this article, we address offline analysis, considering that we are in an exploratory analysis scenario (the detection models are not deployed subsequently). Therefore, in the experimental section, all data will be used for training, and the results will also be presented in a training context.

\subsection{Cross-knowledge}

When it comes to analyzing multiple time series, pooling knowledge allows to extract richer information compared to treating each time series individually. By combining different series, it is possible to detect trends, correlations or patterns that would not be visible when analyzing each time series separately. This approach provides a deeper understanding of the interactions and dynamics between different time series. By exploiting this cross-referenced information, it becomes possible to improve forecast accuracy and identify anomalies or unusual events. However, as mentioned in \cite{data_benchmark_problem}: ``each time series must be considered as totally independent of the others, unless we know the interactions or clearly the application domain in which the time series are generated''. In this article, however, we are as agnostic as possible. As a result, we will be studying time series one by one, without combining information. Each time series thus becomes a ``dataset'' in its own right, on which a detection method can be applied and performance results collected.

\subsection{Typology of Approaches}

The literature considers three main families of anomaly detection approaches \cite{boniol-slides}: 
\begin{itemize}
    \item (i) \textbf{supervised}, where each time series, or a portion of a it, is labeled as either normal or anomalous.
    \item (ii) \textbf{semi-supervised}, where it is assumed that the beginning of the time series contains no anomalies; in this case, the model is trained on the normal data only and then deployed on the rest of the time series for anomaly detection.
    \item (iii) \textbf{unsupervised}, where no assumption is made: anomalies can occur at any point in the time series and and no labels (normal/anomalous) are available during training. 
\end{itemize}

In this article, consistent with the idea of being agnostic and performing cold analysis, we place ourselves in the third case.

\subsection{Detection Methods: Time-Series vs. Tabular} 

There are numerous anomaly detection methods for univariate time series \cite{Braei2020AnomalyDI,6684530} and just as many for tabular data \cite{Chandola2009}. It is interesting to note that these two fields share certain methods, ranging from the simplest (statistical methods, for example) to the most elaborate. Indeed, methods designed for tabular data often perform very well on temporal data. Examples of these include Isolation Forest (IF) \cite{iForest} and Local Outlier Factor (LOF) \cite{lof}. In the oral presentation made by Boniol in \cite{boniol-slides}, the interested reader will find a comparison of a fairly large number of methods (in particular slides 125 to 130), showing the very good positioning of IF and LOF on temporal data. Given the objective of the study presented in this article, and the fact that both methods work well in both the temporal and tabular domains, they both will be the methods we use in the remainder of our comparative study. 

\subsection{Time Series Windowing}

The final concept we need to introduce is ``sliding windows'' (and the associated parameters), which allows methods initially dedicated for tabular data to be used on temporal data. For a given time series (in which we wish to detect anomalies), the method consists in ``slicing'' it into a succession of $F$ windows (of size $W$). These windows are then organized into a table of $W$ columns and $F$ rows, thus turning the time series into a table. The objective is then to identify which rows of this table contain anomalies.

The window passed over the time series can be either ``jumping'' (also known as non-overlapping), or ``sliding'' (overlapping). In the sliding case, an additional parameter is introduced: the shift step $\alpha$ which determines the degree of overlap between consecutive windows (jumping windows can be thought of as sliding windows with $\alpha = W$, i.e. no overlap). A fairly careful review of the literature \cite{TSAD_comprehensive_evaluation,TSAD_survey_SOTA,PA_problem,kapourchali2018unsupervised,shawe2015novelty,golmohammadi2015time,geiger2020tadgan} shows that most methods, unless they incorporate feedback from an expert, use a sliding window and $\alpha = 1$. This will also be our case in the remainder of this article.

The remaining challenge is to determine the window size ($W$). This point is quite crucial, yet surprisingly under-explored in the literature. Most studies set the value of $W$ through preliminary cross-validation experiments. To our knowledge, only one paper has reasonably addressed the question of automatically determining $W$ in the case of ``periodic'' time series \cite{arik}, in which the authors test several methods for detecting ``seasonality/periodicity'', within a semi-supervised framework. In our case, we will not make the assumption that the time series studied are periodic. The experimental section of this article will therefore test several values of $W$ with increasing orders of magnitude.

\subsection{Feature Extraction from Time Series}

The approach involves transforming data from the temporal world to the tabular world using libraries that automatically extract features from time series. The features are diverse, allowing to capture different properties of the time series, such as seasonality, trends or auto-correlation, and can therefore be adapted to different application domains. This transformation thus captures the essence of time series data, while making it compatible with more traditional tabular data analysis techniques.

\begin{figure*}[!ht]
    \centering
    \includegraphics[width=1.0\linewidth]{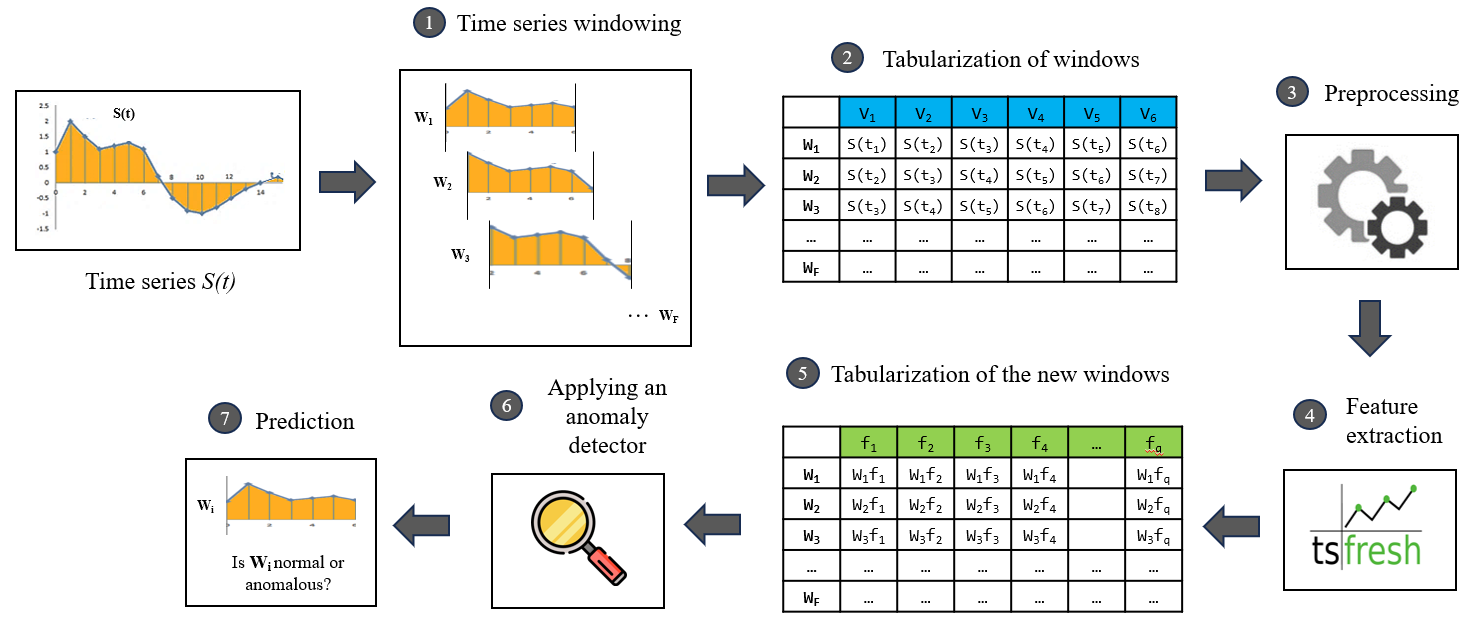} 
    \caption{Proposed processing pipeline.}
    \label{pipeline}
\end{figure*}

Stimulated by the development of the library HCTSA \cite{fulcher2017hctsa}, several unsupervised feature engineering tools have been developed independently in different programming languages: for example, CATCH22 \cite{LSK+19}, FEATURETOOLS \cite{KM15}, TSFRESH \cite{CBN+18}, TSFEL \cite{BFF+20}, TSFEATURES \cite{hkm+23}, or FEASTS \cite{o2020package}. An extensive study presented in \cite{classif} (for the classification task) but also those presented in \cite{features_engineering_DBSCAN,features_engineering_PCA} (for the anomaly detection task) show that TSFRESH \cite{CBN+18} is among the best performers. It is therefore our selected feature extraction tool. The proposed approach could nonetheless use other tools, or even combine them (for example using a stacking method).

TSFRESH \footnote{\url{https://tsfresh.readthedocs.io/en/latest/index.html}} (Time Series FeatuRe Extraction on basis of Scalable Hypothesis tests) is a Python library that calculates up to 800 features (basic statistics, autocorrelations, entropy, Fourier coefficients, etc.) from a time series by combining 63 time series characterization methods \cite{CBN+18}. It offers three pre-defined feature dictionaries ranging in size from 10 to almost 800 features. The library also enables feature selection based on statistical tests. This process can be parallelized, to reduce computation times. In this paper, we used the maximum number of features, as feature selection is only possible in a supervised framework.

\section{Proposed Processing Pipeline}
\label{sec-pipeline}

Based on all the concepts presented previously in this paper, the processing pipeline proposed for our comparative study is presented in figure \ref{pipeline}. It consists of seven steps and is applied to each time series:

\begin{itemize}
\item Step 1: divide the series into windows ($W=6$ for illustration purposes in figure \ref{pipeline});
\item Step 2: put the obtained $F$ windows, of size $W$, ($W_1$ to $W_F$) in a tabular data base;
\item Step 3: preprocess data;
\item Step 4: feature extraction using the TSFRESH library;
\item Step 5: put the $F$ obtained windows ($W_1$ to $W_F$), described by the $q$ features calculated using TSFRESH, in a tabular data base;
\item Step 6: apply an anomaly detector (IF or LOF);
\item Step 7: obtain the anomaly prediction for each window.
\end{itemize}

The experimental study described in the following section will compare the results obtained with and without steps 4 and 5, to answer the research question raised in the introduction of this paper.


\section{Experimental Protocol}
\label{sec-protocole}

This section describes the choices made during the experiments: user-defined parameters, sliding window size ($W$), etc. All experiments carried out can be reproduced using the code in \cite{suplementarymaterial-github}. Some preliminary tests, which will be briefly discussed below, are not fully presented in this article but interested readers can nevertheless find them in the ``Additional material'' file available in the GitHub mentioned above \cite{suplementarymaterial-github}.

\begin{table*}[t]
\centering
\caption{Summary of information on the datasets used.}
\begin{tabular}{|c||c|c|c|c|c|c|c|}
\hline
Name & \#series & \multicolumn{2}{|c|}{Length} & Freq (Hz) & \multicolumn{2}{|c|}{\% anomalies} & Domains \\ \hline
\multicolumn{2}{|c|}{} & Min & Max & & Min & Max & \\ \hline\hline
SVDB &  76 & 230 400 & 230 400 & 0.008  & 0.34 & 45.54 & ECG\\\hline
NAB &  46 & 1 127 & 22 695 & variable & 8.30 & 10.29 & servers, tweets, traffic, advertising\\\hline
AIOPS &  13 & 16 441 & 295 414 & [0.003-1.7]  & 0.06 & 7.50 & performance indicators\\\hline
NormA & 5 & 2 000 & 35 040 & unknown & 3.08 & 9.13 & aerospace, health, body language, electricity\\\hline
UCR &  247 & 6 674 & 300 262 & diverse & 0.0005 & 4.9 & medical, meteorology, biology, industry\\
\hline
\end{tabular}
\label{resume_data}
\end{table*}

\subsection{Datasets}
\label{dataset}

Anomaly detection is currently the subject of a great deal of research, and numerous benchmark datasets have been established to enable comparing the performance of new algorithms to the state of the art. Among these, we have selected a number of datasets described below. In some cases, we did not retain all time series from these datasets. The reasons for excluding certain time series are given, and the exact list of time series (identifiers) used for each dataset is available in  \cite{suplementarymaterial-github}.

{\bf SVDB} (MIT-BIH Supraventricular Arrhythmia Database) (available at \cite{timeeval}): this dataset initially contains 78 time series.
However, two series have been discarded due to an extremely high contamination rate (CT) (>50$\%$) which unrealistic.

{\bf NAB} (Numenta Anomaly Benchmark) (available at \cite{timeeval}): created by Numenta in 2015, it is currently the second most widely used benchmark in the literature \cite{data_benchmark_problem}. It consists of 58 time series divided into seven groups, two of which contain synthetic data only. We chose not to use these synthetic data, which correspond to 11 time series. The remaining five groups cover a variety of topics \footnote{more information at: \url{https://github.com/numenta/NAB/tree/master/data}}). From these five groups, we removed one time series that contains no anomalies. In total, we used 46 time series from this dataset.

{\bf AIOPS 2018}\footnote{Available at: \url{https://github.com/TheDatumOrg/TSB-UAD}}: this dataset was created in 2018 for the AIOps challenge. 29 time series were collected from various companies such as Sogou, Tencent or eBay. They correspond to performance indicators that reflect the scale, quality of web services and health of a machine as explained in \cite{kpi_dataset}. The data we used comes from the benchmark dataset \cite{paparrizos2022tsb}. Additionally, it is worth noting that we have combined the training and testing parts for each time series. After analyzing the correlations among the 29 time series, we noticed that some of them are correlated, which could potentially introduce bias in the statistical tests conducted during the analysis of the results. Therefore, we only keep the uncorrelated series ($r < 0.3$), ultimately leading us to select 13 series for this dataset.

{\bf NormA} (available at \cite{timeeval}): this dataset consists of 21 time series, of which 14 are synthetic and have been discarded. The seven real series can be grouped into four categories. Among the real-world time series, the first three are highly correlated, and we will thus only keep one of the three. In total, we thus keep five time series only for this dataset.

{\bf UCR Anomaly Archive}\footnote{Available at: \url{https://www.cs.ucr.edu/~eamonn/time_series_data_2018/}} : The UCR Anomaly Archive dataset was released in 2020 as an alternative to several benchmarks deemed to be deficient by \cite{data_benchmark_problem}.
For this dataset, we chose to only exclude three series that are significantly longer than the others (due to computational time constraints). We thus used a total of 247 time series from this dataset.

To sum up, our five datasets\footnote{Norma and NAB have sometimes described having mislabeled ground truth, trivial or overestimated success by having repeated anomalies. But in our case, as mentioned above, by removing synthetic data, correlated time series,
and manual inspection we have done our best to avoid these aspects for these two datasets (see \cite{suplementarymaterial-github} to have the exact id of the retained time series).} together represent a wide diversity of problems, anomaly rates and issues
(see Table \ref{resume_data}). When presenting the results, the AIOPS and NormA datasets, which contain a small number of time series, will be merged in order to perform statistical tests, such as the Wilcoxon test \cite{Wilcoxon1992}, which requires a minimum number of values to be used.

\subsection{Size of Sliding Windows}
\label{fenetre}
As mentioned above, this point is quite crucial but surprisingly little discussed in the literature. To the best of our knowledge, only one paper has reasonably addressed the question of automatically adjusting the value of $W$ in the case of ``periodic'’ time series \cite{arik}. In our case, where we aim to remain agnostic and without feedback loops, we decided to simply test four window sizes corresponding to values observed in the literature. We started with a small size $W=32$ and then doubled its size, hoping to increase the amount of information captured, several times. In the end, the tested values for $W$ are 32, 64, 128, and 256. The aim of the experiments will not be to determine the optimal window size but to confirm that the conclusions do not depend on a very specific window size.

\subsection{Hyperparameter Tuning}
\label{tsfresh}

The TSFRESH library \cite{CBN+18} (acronym for \textit{Time Series FeatuRe Extraction on basis of Scalable Hypothesis tests}) can be used to create varying numbers of features (the value of $q$ mentioned in section \ref{sec-pipeline}). All the tests we conducted showed better results by pushing the feature extraction process to its limits, thus using a large number of features calculated by the library (i.e. its ‘Efficient’ version and $q=777$ features). We point out that sometimes, for example with $W=32$, it is not possible for TSFRESH to calculate all the requested features. In such cases, we removed the impacted columns (with missing values) from the table produced in step 5 of the pipeline described in section \ref{sec-pipeline}.

Two anomaly detectors for tabular data that have shown their worth on time series are used: Isolation Forest (IF) and Local Outlier Factor (LOF). In all the experiments carried out below, we used the implementation provided in the PyOD\footnote{\url{https://pyod.readthedocs.io/en/latest/}} library (version 1.1.3) with their default settings.

\subsection{Results Evaluation Criterion}

The literature offers numerous evaluation metrics, such as accuracy, F1 score, the ``PA\%K'' criterion \cite{Kim2021TowardsAR}, the AUC (\textit{Area Under the receiver operating characteristics Curve}) \cite{FAWCETT2006}. In our case, we decided not to use a criterion that requires a threshold value, as we place ourselves in an agnostic framework where there is no feedback loop with a domain expert. Indeed, this threshold can be a difficult parameter to select and/or may require starting from a heuristic on the distribution of anomaly scores. For this reason, we selected the AUC as our evaluation criterion.

The reader will note that in this case, but also for other potential criteria, ground truth labels are necessary to allow the evaluation of the predictions obtained in step 7 of our processing pipeline. For the predicted class, this information is given by the detection method used, here IF or LOF. For the class to be predicted (ground truth), we operate as follows (which corresponds to the observed usage): during the segmentation of the time series into windows (step 1), if an anomaly exists in the window, then the entire window is labeled as anomalous, and vice versa in the opposite case. 
Thus, it is possible to compare for each window the class predicted by a confidence score delivered by IF or LOF with the ground truth. The set of comparisons allows us to calculate the AUC for each time series (individually) from the datasets used. The collection of these AUCs will also enable us to conduct statistical tests and present results in terms of ranking or critical diagrams.

\subsection{Preprocessing}

Preliminary tests were carried out in order to analyze the benefits from ``horizontally normalizing'' the values contained in the sliding windows before applying the two approaches: 1) anomaly detection using the initial representation (called ``TS'' in Table \ref{toto}) and 2) anomaly detection using the new feature representation calculated by TSFRESH (called ``FE'' in Table \ref{toto}). To this end, a comparative study was carried out between ``doing nothing'' (No Normalization) and 3 usual normalization methods, namely (i) Min-Max, (ii) Median-IQR and (iii) Mean-Standard Deviation.

This study is conducted based on (i) three of the datasets described above: SVDB, NormA and AIOPS among the five selected and (ii) the TSFRESH library parameter set to ‘minimal’ ($q=10$). Since the NormA dataset only consists of five time series, its results are combined with those of AIOPS, as mentioned earlier.

\begin{table}[t]
\centering
\setlength{\tabcolsep}{4pt} 
\caption{Comparison of the mean ranks of normalization approaches by window size (combining IF and LOF results).}
\begin{tabular}{|p{1.1cm}|c|c||c|c|c|c|}
\hline
 &  & &  {Without} & {} & {} & {} \\ 
 &  & &  {Normalization} & {MinMax} & {Median-IQR} & {MeanStd} \\ 
 \hline
\hline
\multirow{8}{*}{{SVDB}} & \multirow{2}{*}{{32}} & {TS} & 2.250 & 2.036 & 3.063 & 2.651 \\ 
 &  & {FE} & 1.474 & 2.250 & 3.135 & 3.141 \\ 
\cline{2-7}
 & \multirow{2}{*}{{64}} & {TS} & 2.211 & 2.227  & 2.908 & 2.655\\ 
 &  & {FE} & 1.750 & 2.260  & 2.934 & 3.056\\ 
 \cline{2-7}
 & \multirow{2}{*}{{128}} & {TS} & 2.174 & 2.243 & 2.609 &  2.974 \\ 
 &  & {FE} & 2.049 & 2.418 & 2.474 & 3.059 \\ 
\cline{2-7}
 & \multirow{2}{*}{{256}} & {TS} & 2.079 & 2.638 & 2.424 & 2.859 \\ 
 &  & {FE} & 2.260 & 2.563 & 2.230 & 2.947 \\ 
\hline
\hline
\multirow{8}{=}{{AIOPS + NormA}} & \multirow{2}{*}{{32}} & {TS} & 1.806 & 2.583 & 2.472 & 3.139 \\ 
 &  & {FE} & 2.236 & 2.708 & 2.750 & 2.306\\ 
\cline{2-7}
 & \multirow{2}{*}{{64}} & {TS} & 1.903 & 2.431 & 2.722 &  2.944 \\ 
 &  & {FE} &  2.458 &  2.722 & 2.625 & 2.194 \\ 
 \cline{2-7}
 & \multirow{2}{*}{{128}} & {TS} &  1.722 & 2.583 & 2.667 & 3.028\\ 
 &  & {FE} & 2.500 &2.722 & 2.417 & 2.361\\ 
 \cline{2-7}
 & \multirow{2}{*}{{256}} & {TS} & 1.736 &  2.944 & 2.319 & 3.000 \\ 
 &  & {FE} & 2.250 &  2.944 &  2.250 & 2.556 \\ 
\hline
\end{tabular}
\label{toto}
\end{table}

The results presented in Table \ref{toto} show the mean rank of these four normalization techniques (1 being the best, 4 the worst), aggregated regardless of the detection method (IF or LOF) and versus the window size. We see that normalization approaches do not improve performance. The absence of normalization exhibits the best mean rank in most experiments, and thus yields the best results. Consequently, in the next section, only the results obtained with no normalization are presented. A more in-depth analysis of this study found in the supplementary material \cite{suplementarymaterial-github}, according to the “detection method” axis (IF or LOF), further confirms this result.


\section{Results}
\label{sec-resultats}

\subsection{Performance Gains from the Extracted Features for Each Detector}

Table \ref{table_IF_et_LOF_efficient} reports the obtained results: with (FE) or without (TS) the use of TSFRESH in the proposed processing pipeline; with the Isolation Forest (IF) or Local Outlier Factor (LOF) detectors. These results are detailed by dataset and window size. The presented values correspond to the mean ranks obtained over all time series of each dataset. The p-values are derived from the Wilcoxon test \cite{Wilcoxon1992}: a bold value indicates a statistically significant difference in performance between the pair of detectors.

\begin{table*}[t]
\centering
\caption{Comparison of detectors, by dataset, according to the window size used.}
\begin{tabular}{|c|c||c|c|c||c|c|c|}
\hline

& & \multicolumn{3}{|c|}{Isolation Forest} & \multicolumn{3}{|c|}{Local Outlier Factor} \\
\hline
\hline
 &  &   {TS} & {FE} & {p-value}   &   {TS} & {FE} & {p-value} \\ 
\hline
\hline
\multirow{4}{*}{{SVDB}} & {32} & 1.882 & 1.118   & {\bf 1.522$\times \text{10}^{\text{-12}}$} & 1.329 & 1.671   & {\bf 2.066$\times \text{10}^{\text{-3}}$}\\ 
 &  {64}& 1.855 &  1.145 &   {\bf 5.375$\times \text{10}^{\text{-9}}$} & 1.480 & 1.520  &  { 4.454$\times \text{10}^{\text{-1}}$}\\ 
 &  {128} & 1.842 & 1.158 &   {\bf 5.553$\times \text{10}^{\text{-10}}$} & 1.750 & 1.250  &   {\bf 2.490$\times \text{10}^{\text{-6}}$}\\ 
 &  {256} & 1.592 & 1.408  &  1.161$\times 10^{-1}$ & 1.684 & 1.316  &  {\bf 5.541$\times\text{10}^{\text{-6}}$} \\ 
\hline
\hline
\multirow{4}{*}{{AIOPS + NormA}} & {32}   & 1.889 & 1.111 & {\bf 1.930$\times \text{10}^{\text{-3}}$} & 1.556 & 1.444  & { 8.317$\times \text{10}^{\text{-1}}$}\\ 
 &  {64} & 1.778 & 1.222  & {\bf 1.930$\times\text{10}^{\text{-3}}$} & 1.556 & 1.444 & { 7.660$\times\text{10}^{\text{-1}}$} \\ 
 &  {128} & 1.778 & 1.222  & {\bf 1.930$\times \text{10}^{\text{-3}}$} & 1.417 & 1.583  & { 4.925$\times \text{10}^{\text{-1}}$} \\ 
 &  {256}  & 1.833 & 1.167 & {\bf 4.745$\times \text{10}^{\text{-3}}$} & 1.389 & 1.611 & {1.674$\times \text{10}^{\text{-1}}$}\\ 
\hline
\hline
\multirow{4}{*}{{NAB}} & {32} & 1.652 & 1.348  & 5.059$\times 10^{-2}$ & 1.261  & 1.739 & {\bf 8.504$\times \text{10}^{\text{-4}}$}\\ 
 &  {64}& 1.630 & 1.370  & {\bf 1.727$\times \text{10}^{\text{-2}}$} & 1.261 & 1.739  & {\bf 7.430$\times \text{10}^{\text{-5}}$} \\ 
 &  {128} & 1.674  & 1.326 &  {\bf 5.525$\times \text{10}^{\text{-3}}$} & 1.217 & 1.783 &  {\bf 1.261$\times \text{10}^{\text{-4}}$} \\ 
 &  {256} & 1.641 & 1.359  & {\bf 1.723$\times \text{10}^{\text{-2}}$} & 1.283 & 1.717  & {\bf 3.163$\times \text{10}^{\text{-3}}$}\\ 
\hline
\hline
\multirow{4}{*}{{UCR}} & {32} & 1.826 & 1.174  & {\bf 2.581$\times \text{10}^{\text{-29}}$} & 1.156 & 1.844  & {\bf 5.159$\times \text{10}^{\text{-22}}$} \\ 
 &  {64} & 1.844 & 1.156  & {\bf 9.264$\times \text{10}^{\text{-32}}$} & 1.170 & 1.830  & {\bf 7.227$\times \text{10}^{\text{-26}}$} \\ 
 &  {128}  & 1.848 & 1.152 &  {\bf 1.792$\times \text{10}^{\text{-32}}$}  & 1.223 & 1.777  &  {\bf 2.247$\times \text{10}^{\text{-20}}$} \\ 
 &  {256} & 1.796 & 1.204  & {\bf 8.983$\times \text{10}^{\text{-27}}$} & 1.243 & 1.757  & {\bf 1.055$\times \text{10}^{\text{-19}}$} \\ 
\hline
\end{tabular}
\label{table_IF_et_LOF_efficient}
\end{table*}

The results are clear: (i) for Isolation Forest, across 16 results (4 datasets x 4 window sizes), the proposed FE method (using TSFRESH) always obtains the best mean rank with 14/16 cases where the difference is statistically significant (ii) for Local Outlier Detection, the proposed FE method obtains the best mean rank 4 times, of which only 2 are significantly better. The proposed processing chain benefits IF, which is tree-based (thus relying on value ranks), but does not benefits LOF, which is density-based (thus relying on distance calculations).

It is likely that for LOF, the size of the vector produced by TSFRESH leads to a dimensional problem in the distance calculations used by this method, whereas IF does not suffer from this problem. The performance of LOF nevertheless raises questions. This is a model that requires more attention than Isolation Forest (based on rank statistics), which needed neither normalization nor optimization of its hyperparameters. It is also conceivable that adjusting the value of $k$ in LOF could provide an improvement, but in the case without a feedback loop with a domain expert, we have not explored this possibility. These points will be explored in a future work. A normalization of the extracted features (vertically) was also evaluated (as an additional step between the step 5 and 6 in Figure \ref{pipeline}) the results indicate that: 1) for UCR and (AIOPS + NormA) datasets, LOF-FE nearly always improved its performance but it was not enough to surpass LOF-TS, 2) for the NAB dataset, LOF-FE consistently improved its performance and surpassed LOF-TS and 3) for SVDB, a decline in the performance of LOF-FE was observed. Our conclusion remains therefore unchanged.

\subsection{Comparison of detectors}

\begin{figure*}[!ht]
    \centering
    \includegraphics[width =1.0\linewidth]{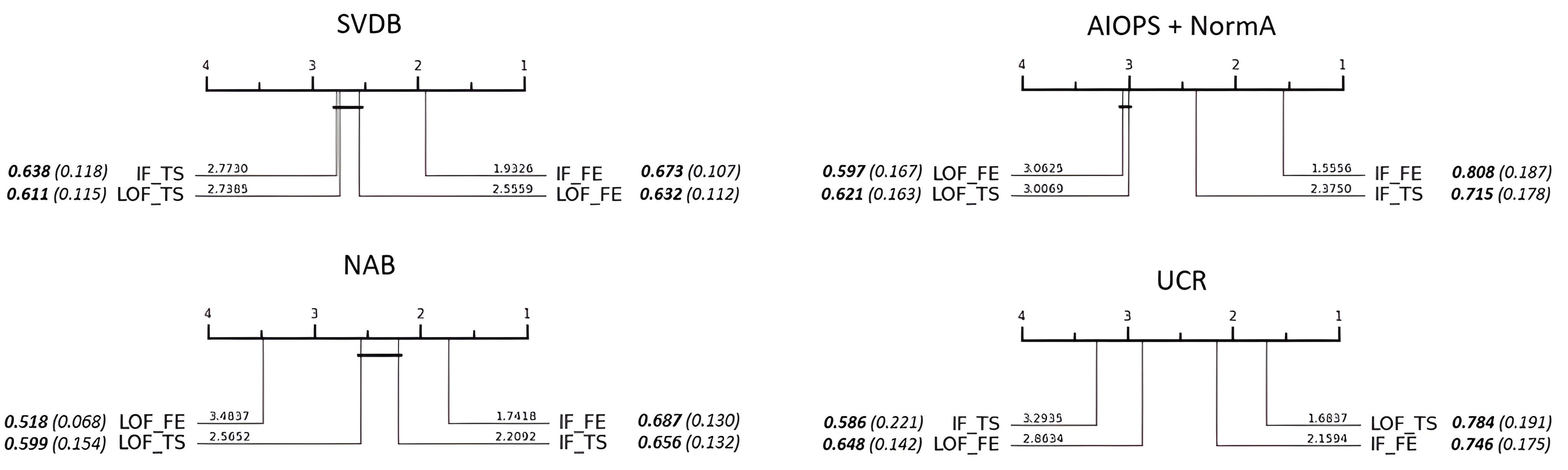} 
    \caption{Critical diagrams for each dataset, averaged over the four window sizes.}
    \label{article_CDD_IF_LOF}
\end{figure*}

Figure \ref{article_CDD_IF_LOF} shows the critical diagram (CD) of methods (i) per dataset, then (ii) with or without the use of calculated features and (iii) averaged over all window sizes. In each sub-figure and for each bar of the critical diagram, we find in this order: the mean rank of the method, the method name (IF or LOF) prefixed with TS or FE (respectively for initial Time Series or FE for Feature Engineering via TSFRESH), then the corresponding mean AUC value with finally the standard deviation of the AUC in brackets. Interested readers will find further results with other axes of analysis in the supplementary material \cite{suplementarymaterial-github}.

Once again, we note that, in general, Isolation Forest benefits most from the features calculated using TSFRESH, ranking first on 3 of the 4 datasets. For LOF, the results are much more contrasted: the creation of features brings little or no improvement compared to the initial temporal representation. These results confirm those presented in Table \ref{table_IF_et_LOF_efficient}. Finally, LOF applied to temporal data achieves the best results on the UCR dataset. This dataset has the particularity of containing very few anomalies (see Section \ref{dataset}), with anomalies appearing only in the second part of each time series. This has a remarkable impact on IF which dramatically improves its performance using features of TSFRESH: AUC moves from 0.586 to 0.746, i.e. +36\%.


\section{Conclusion}
\label{sec-conclusion}

In the context of anomaly detection within a time series, this paper proposed a processing pipeline which first transforms time series from the temporal domain to an alternative tabular representation space, and then apply anomaly detectors dedicated to tabular data. Our motivation is to consider the feature extraction process as a source of knowledge and information useful for detection. The central question was whether better results could be obtained in the computed feature space compared to the initial temporal representation? Through extensive experiments across five datasets and with two detectors (IF and LOF), we observed that the results significantly improved for IF, but not for LOF. Future work will focus on (i) extending the number of detectors in the comparison, (ii) testing other feature extraction libraries (or combining them), and (iii) considering tuning the window size and/or considering not a sliding window but a ``jumping'' window (with no overlapping between consecutive windows), even though this is not inherent to the proposed processing chain.

\bibliographystyle{IEEEtran}
\bibliography{references}

\end{document}